\pdfoutput=1

\documentclass[11pt]{article}

\usepackage[final]{acl}

\usepackage{times}
\usepackage{latexsym}

\usepackage[T1]{fontenc}

\usepackage[utf8]{inputenc}

\usepackage{microtype}

\usepackage{inconsolata}

\usepackage{graphicx}

\usepackage{amsmath, amsthm, amssymb}
\usepackage{xspace}

\usepackage{cleveref}

\usepackage{paralist}

\usepackage{booktabs}
\usepackage{multirow}
\usepackage{amsmath} 
\usepackage{amssymb}
\usepackage{tikz}
\usetikzlibrary{positioning}
\usetikzlibrary{decorations.pathmorphing, positioning}

\usepackage{graphicx} 
\usepackage{xcolor}   
\usepackage{listings} 

\usepackage{mdframed}

\lstdefinelanguage{SQL}{
    keywords={SELECT, FROM, WHERE, AND, OR, AS, COUNT, JOIN, ON, GROUP BY, WITH, MAX},
    keywordstyle=\bfseries\color{blue},
    morekeywords={[2]{Medal, Format, Tournament, Athlete}},
    keywordstyle={[2]\color{teal}},
    sensitive=true,
    comment=[l]{--},
    morecomment=[s]{/*}{*/},
    commentstyle=\color{gray},
    stringstyle=\color{purple},
    morestring=[b]',
    literate={á}{{\'a}}1 {Á}{{\'A}}1 {é}{{\'e}}1 {É}{{\'E}}1 {í}{{\'i}}1 {Í}{{\'I}}1 {ú}{{\'u}}1 {Ú}{{\'U}}1 {ó}{{\'o}}1 {Ó}{{\'O}}1 {ç}{{\c{c}}}1 {Ç}{{\c{C}}}1,
}

\definecolor{orange}{rgb}{1,0.5,0}
\definecolor{mdgreen}{rgb}{0.05,0.6,0.05}
\definecolor{mdblue}{rgb}{0,0,0.7}
\definecolor{dkblue}{rgb}{0,0,0.5}
\definecolor{dkgray}{rgb}{0.3,0.3,0.3}
\definecolor{slate}{rgb}{0.25,0.25,0.4}
\definecolor{gray}{rgb}{0.5,0.5,0.5}
\definecolor{ltgray}{rgb}{0.7,0.7,0.7}
\definecolor{purple}{rgb}{0.7,0,1.0}
\definecolor{lavender}{rgb}{0.65,0.55,1.0}

\definecolor{darkgreen}{RGB}{0,150,0}
\definecolor{darkred}{RGB}{250,0,0}
\definecolor{lightgray}{RGB}{240,240,240}

\newcommand{\concept}[1]{`\emph{#1}'}

\title{Reinforcing Code Generation: Improving Text-to-SQL with Execution-Based Learning
}

\author{
  \textbf{Atharv Kulkarni}\textsuperscript{1} \quad
  \textbf{Vivek Srikumar}\textsuperscript{1} \\
  \textsuperscript{1}University of Utah \\
  {\tt \small u1322897@utah.edu \quad svivek@cs.utah.edu}
}

\begin{document}
\maketitle

\begin{abstract}
In this work, we study the problem of code generation with a large language model (LLM), with a focus on generating SQL queries from natural language questions. We ask: Instead of using supervised fine tuning with text-code pairs, can we tune a model by having it interact with a database engine?

We frame this problem as a reinforcement learning problem where the model receives execution-based feedback from the environment in the form of scalar rewards. These rewards penalize execution failures and assign positive values when a query returns a correct answer. We use the rewards within the Group Relative Policy Optimization (GRPO) framework. 

We use a tabular reasoning benchmark to test and evaluate our findings. We find that with only weak supervision in the form of question-answer pairs, RL-tuning improves the accuracy of model generated SQL code from 31.49 to 49.83 while reducing error percentage from 25.43\% to 14.71\%. This improvement allowed the model nearly match the performance performance to the larger SQLCoder-70B model. Our work demonstrates the potential of using  execution-based feedback to improve symbolic reasoning capabilities of LLMs.

\begin{figure}[t]
    \centering
    \includegraphics[width=0.5\textwidth]{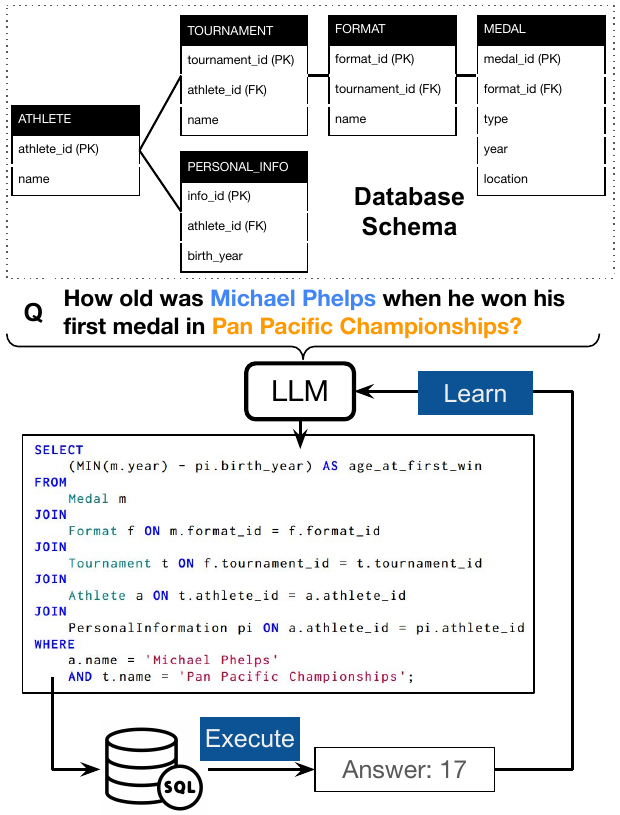}
    \caption{Given a NLQ and database schema, the LLM generates a SQL query that when executed on the database returns the answer. The answer helps the LLM improve its code generation ability.}
    \label{fig:pipeline-example}
\end{figure}
\end{abstract}

\section{Introduction}
\label{sec:intro}

Large language models (LLMs) have enabled natural language processing models to understand and generate sophisticated outputs across diverse tasks~\cite[e.g.,][]{DBLP:journals/corr/abs-2005-14165, grattafiori2024llama3herdmodels}. Building on these successes, recent focus has shifted towards using these them as agentic AI systems, capable of actively interacting with external environments and tools~\cite{zhou2024webarenarealisticwebenvironment, wang2024executablecodeactionselicit}. The agentic paradigm treats LLMs as autonomous entities that execute actions, interpret environmental feedback and iteratively refine their behavior through direct interaction~\cite{chen2021evaluatinglargelanguagemodels, austin2021programsynthesislargelanguage}.

Developing agentic AI requires capabilities beyond mere generative tasks, particularly involving symbolic reasoning. One important capability is the ability to generate executable code in a target programming language (e.g., Python, SQL, etc.) from natural language inputs.
Accurate code generation can allow agentic models to actively interact with structured data and make decisions based on the insights gained. This capability transforms static generative models into interactive agents capable of reasoning about structured sources such as databases, APIs, webpages and spreadsheets.

Traditionally, training models for code generation has relied on supervised fine tuning (SFT), which requires large labeled datasets that map text to code~\cite{codegemmateam2024codegemmaopencodemodels, srivastava2023sqlcoder}. These may be available via public software repositories; yet, merely supervised fine-tuning will not endow the models with the ability to easily generalize to new kinds of execution environments that may involve new database schemas or APIs. To do so, we need the models to engage with these  environments and elicit feedback from their behavior. This naturally calls for using reinforcement learning (RL) with an interpreter in the loop.

In this paper, we focus on the specific programming problem of generating SQL queries given a database schema and a natural language question.  We investigate reinforcement learning as a method to generate accurate SQL queries (see~\cref{fig:pipeline-example}). We specifically explore three research questions:

\begin{itemize}
\item \textbf{Q1:} Can RL-trained models outperform specialized SQL generation models in terms of answer quality and syntactic validity?
\item \textbf{Q2:} How does RL training affect performance across queries of varying complexity?
\item \textbf{Q3:} Can RL training significantly enhance robustness on counterfactual temporal data?
\end{itemize}

We evaluate our approach using the TEMPTABQA-C benchmark~\cite{kulkarni-etal-2025-llm}, which consists of natural language queries about temporal information in Wikipedia infoboxes. Our results show that RL-trained agents achieve higher accuracy and robustness compared to traditional supervised models, particularly in handling complex queries. Moreover, with counterfactual inputs (representing unseen kinds of data), we find that after RL training, a 7B SQLCoder model performs the SQLCoder with 70B parameters . Our findings demonstrate the merits of execution-based reinforcement learning in developing robust agentic AI systems.\footnote{The code for replicating our experiments is forthcoming.}

\section{Problem Statement}
\label{sec:problem}

Recent LLMs have demonstrated the potential to serve as autonomous agents that can complete complex tasks \cite[e.g.,][]{zhou2024webarenarealisticwebenvironment}. A key component of this capability is the ability to interact with external tools and systems, such as databases, web interfaces, and APIs. To facilitate such interaction, models can generate code ~\cite[e.g.][]{wang2024executablecodeactionselicit}, allowing them to communicate precise intentions, invoke tool-specific functionalities, and obtain deterministic outputs. This strategy aligns with the idea of tool-augmented reasoning, where models generate executable code (in SQL, Python, etc.) to access or manipulate external environments in a structured and verifiable manner~\cite{DBLP:journals/corr/abs-2107-03374, austin2021programsynthesislargelanguage}.

A technical difficulty with the above strategy is that for each task, program or API, we must train language models that understand the interface and produce valid outputs in the corresponding formal language. The standard approach is supervised fine-tuning (SFT), where models are trained on text-to-program mappings to learn the desired mapping. SFT has sown strong performance in code generated settings, as demonstrated by open-weights models like CodeGemma~\cite{codegemmateam2024codegemmaopencodemodels} and SQLCoder~\cite{srivastava2023sqlcoder}.

SFT requires large amounts of high-quality labeled data and significant computational resources, neither of which may be available for new programming languages, APIs or symbolic environments. If the models were trained on limited supervision, they may fail to generalize to new inputs. These limitations motivate the need for alternative training paradigms that can learn from weaker forms of feedback, like execution outcomes, without relying on large-scale code annotation.

We consider an alternative protocol for fine-tuning models to write code. Instead of supervising with provided code, we treat the model as an agent interacting with a database engine in a given environment to solve a structured reasoning task. Specifically, we focus on an agent that interacts with a database, provided with a natural language query (NLQ) and a database \emph{schema} describing table structure and relationships. The model's goal is to generate an SQL query that answers the question when executed on the database, learning through interaction with the engine. This setup is analogous to semantic parsing with world feedback, where models learn from their interactions with a physical or simulated environment~\cite[e.g.,][]{clarke-etal-2010-driving,liang-etal-2011-learning,berant-etal-2013-semantic}.

Learning from the answer supervision requires the model to connect language understanding with symbolic reasoning. It must interpret questions, align them with the schema, and generate a logically valid SQL query by applying reasoning over relational structures in the database schema. This involves selecting the appropriate operations and composing a query that satisfies the intent of the question. As the generated program is executed by an external tool (here, the database engine), it ensures that the model's reasoning is both interpretable and grounded in reality.

In this work, we ask: \emph{Can an LLM improve its program generation ability by interacting with a database environment and learning from the results of its execution?} We argue that doing so allows for a more scalable supervision: from entire programs to mere execution-based feedback. \Cref{fig:pipeline-example} illustrates this setup, showing how the model interacts with the schema, question, and database to generate and execute a SQL query, which can then provide feedback to the model.

%%% Local Variables:
%%% mode: LaTeX
%%% TeX-master: "main"
%%% End:

\section{Learning to Code without Code}
\label{sec:approach}

We investigate if an LLM can improve its ability to generate executable SQL programs by interacting with a formal environment (a database engine), rather than simply observing code. Program correctness is determined solely by execution: we assume that if the answer is correct, so is the code.

In the absence of a gradient signal via the database engine, this becomes a natural reinforcement learning problem. The model acts as an agent, generating SQL programs as actions in response to questions and schemas as input states. Execution of the generated query by the database engine provides the scalar rewards that guide the model's training. The reward signal is the only feedback used to train the model, replacing traditional supervised fine-tuning. 

The rest of this section defines our RL setup and the scalar rewards to guide the LLMs.

\subsection{Group Relative Policy Optimization}
\label{sec:grpo]}

We adopt Group Relative Policy Optimization~\cite[GRPO,][]{shao2024deepseekmathpushinglimitsmathematical},  a recently proposed RL algorithm that has been shown to be especially helpful for  tasks involving symbolic reasoning, e.g., with the  DeepSeek-R1 family of models~\cite{deepseekai2025deepseekr1incentivizingreasoningcapability}. Designed to improve upon  the popular Proximal Policy Optimization~\cite{schulman2017proximalpolicyoptimizationalgorithms} approach, GRPO eliminates the need for a learned value network by directly comparing rewards across multiple completions for each prompt.

For each prompt, the model generates a group of $k$ candidate SQL completions using beam search decoding. Each completion is executed on a database. The outputs are scored using a deterministic reward function (described below), and the rewards are used to compute a normalized advantage $A_i$ for the $i^{th}$ sample in the group as:
\begin{equation}
A_{i} = \frac{r_{i} - \mu_{r}}{\sigma_{r}}
\end{equation}

Here,the term $r_{i}$ is the reward of the $i^{th}$ completion, and $\mu_{r}$, $\sigma_{r}$ are the mean and standard deviation of the group’s rewards respectively. Normalizing the rewards ensures relative feedback rather than absolute reward signals, allowing the model to learn more reliably.

To prevent the model from drifting too far from its pre-trained behavior, a KL divergence penalty is added to the training objective. The  GRPO loss is:

\begin{equation}
\mathcal{L}(\theta) = \mathbb{E}_{y \sim \pi_{\theta}}\left[A(y) - \beta \cdot \mathrm{KL}\bigl(\pi_{\theta}(y) \parallel \pi_{0}(y)\bigr)\right]\label{eq:loss}
\end{equation}

Here, $\pi_{\theta}$ is the current model policy (i.e. the parameters of the LLM we are training), $\pi_{0}$ is the reference (pre-trained) model, and $\beta$ controls the strength of the regularization. 

GRPO is well-suited to our setting because it requires only static rewards and avoids unstable critic learning. It is also efficient to implement and train, making it viable even under modest resource constraints. Perhaps most importantly, the use of execution-based rewards aligns closely with the structure of symbolic tasks like SQL generation, where feedback can be deterministically computed.

\subsection{Rewarding Good Programs}
\label{sec:reward-engineering}

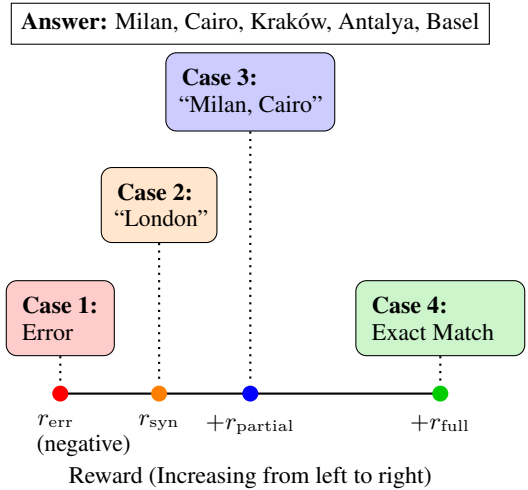
\begin{figure}[tbp]
\begin{center}
\begin{tikzpicture}[x=1cm, y=1cm,
    box/.style={draw, fill=#1!20, rounded corners, align=center, inner sep=6pt},
    dot/.style={fill,#1, circle, inner sep=0pt, minimum size=6pt},
    zigzag/.style={decorate, decoration={zigzag,amplitude=1pt,segment length=4pt}}]

  \node[box=red]  (c1) at (0,1) {\begin{minipage}{1cm}
      \footnotesize \textbf{Case 1:}\\Error
  \end{minipage}};
  
  \node[box=orange] (c2) at (1.3,2.5) {\begin{minipage}{1.1cm}
      \footnotesize \textbf{Case 2:}\\``London''
  \end{minipage}};
  
  \node[box=blue] (c3) at (2.5,4) {\begin{minipage}{1.8cm}
      \footnotesize \textbf{Case 3:}\\``Milan, Cairo''
  \end{minipage}};
  
  \node[box=green!80!black] (c4) at (5,1) {\begin{minipage}{1.8cm}
        \footnotesize \textbf{Case 4:}\\Exact Match
  \end{minipage}};

  \draw[dotted, thick] (c1.south) -- (0,0.15);
  \draw[dotted, thick] (c2.south) -- (1.3,0.15);
  \draw[dotted, thick] (c3.south) -- (2.5,0.15);
  \draw[dotted, thick] (c4.south) -- (5,0.15);

  \draw[thick]
    (0,0) -- (5,0);

  \node[dot=red]    at (0,0) {};
  \node[dot=orange]   at (1.3,0) {};
  \node[dot=blue] at (2.5,0) {};
  \node[dot=green!80!black]  at (5,0) {};

  \node[below=4pt] at (0.2,0) {\begin{minipage}{1cm}
      \footnotesize $r_{\mathrm{err}}$ \\(negative)
  \end{minipage}};
  \node[below=4pt] at (1.3,0) {\footnotesize $r_{\mathrm{syn}}$};
  \node[below=4pt] at (2.5,0) {\footnotesize $ + r_{\mathrm{partial}}$};
  \node[below=4pt] at (5,0) {\footnotesize $+ r_{\mathrm{full}}$};

  \node[below=18pt] at (2.5,-0.2) {\small Reward (Increasing from left to right)};

  \node[above=0.1cm of c3, draw] (a) {\small \textbf{Answer:} Milan, Cairo, Kraków, Antalya, Basel};
  \node[above=0.1cm of a] (q) {\begin{minipage}{3in}
  \small\textbf{Question:} List all the distinct cities where Áron Szilágyi won medals in tournaments held after the year 2021.
  \end{minipage}};
  
\end{tikzpicture}
\caption{This figure illustrates the four kinds of rewards we have. For the given question and the correct answer (a list of five cities), syntax errors incur a negative reward $r_{\mathrm{err}}$. In the absence of an exception, even erroneous answers get a small positive reward $r_{\mathrm{syn}}$. Partially correct answers (case 3) accrue an additional reward $r_{\mathrm{partial}}$, and fully correct answers (case 4) get an additional large reward $r_{\mathrm{full}}$.}
\label{fig:reward_function}
\end{center}
\end{figure}

Accuracy is a natural reward: a policy is given a reward if the result of the SQL it produces exactly matches the ground truth answer, and is given no reward otherwise. However, such a reward tends to be too sparse to provide useful feedback. We construct a reward function that strongly discourages syntactically invalid programs, weakly encourages at least partially correct results, and strongly encourages perfect outputs. To this end, it is comprised of multiple components that reflect these different aspects. See \Cref{fig:reward_function} for an illustration.

In general, the correct answer to a question may be a set (more precisely, a list of rows produced by an SQL query). Our rewards are structured to account for this.

\paragraph{Syntactic Correctness.} If the generated SQL query fails to execute (due to invalid syntax or unresolved references), a penalty of $r_{\text{err}}$ (a negative value) is assigned. Otherwise, the model receives a base reward of $r_{\text{syn}}$. This ensures that invalid programs are strongly penalized while minimal credit is given for queries that merely execute.

\paragraph{Partial Correctness.} If the generated SQL query executes successfully but returns only part of the required answer, we give it a reward $r_{\text{partial}}$. For this purpose, we use the  Relaxed Exact Match Score (REMS), which measures the overlap between the model’s result set and the true result set. Using $r_{\text{partial}}=\text{REMS}$ provides a proportional signal: the model earns more credit as it retrieves more correct items. \Cref{sec:metrics} elaborates on the this score.

\paragraph{Full Correctness.} When the generated SQL query executes without error and returns exactly the ground-truth result set, we assign a reward $r_{\text{full}}$ that is much larger than the partial correctness reward. This exact-match bonus strongly encourages the model to produce fully accurate SQL programs.

The total reward $r$ of a generated output is defined to be the sum of these components. 

\section{Data  \& Experimental Setup}
\label{sec:exp-setup-data}

This section describes the dataset we use for evaluating our RL-based tuning
strategy, and the setup of our experiments.

\subsection{Dataset: TEMPTABQA-C}

Our experiments use the TEMPTABQA-C dataset~\cite{kulkarni-etal-2025-llm}, specifically developed to evaluate and benchmarking LLMs on temporal reasoning tasks with structured relational data. It emphasizes complex temporal reasoning scenarios, counterfactual perturbations, and varied query complexity. Here, we briefly describe the dataset and its components.

The TEMPTABQA-C dataset was generated semi-automatically from Wikipedia infoboxes with temporal information, which were first converted into structured schemas. The resulting schema consists of interdependent tables for athletes, their personal information, tournaments, formats, and medals. These tables form the basis of question-answer pairs that were synthetically generated with templates for the natural language question (NLQ) and an associated SQL query whose execution produces the correct answer. Importantly, in this work, we did not use the SQL queries for training or evaluation and rely solely on the answers.

The questions in the dataset are categorized in terms of their complexity as easy, medium or hard. 

\paragraph{Counterfactual Perturbations.}
To assess robustness, the dataset also includes systematically
generated counterfactual scenarios, created, e.g., by  altering years, changing medal
types, etc. These modified tables maintaining schema consistency and relational
integrity. The original SQL queries, when executed against these altered tables,
produce valid answers under these counterfactual conditions. The counterfactual
subset of the data serves to challenge models to adapt their reasoning rather
than relying on memorized data patterns.

\paragraph{Data Splits \& Statistics.}  Our training data comprises of a broad range of synthetically generated  examples covering different reasoning types. This set consists of 2961 examples with an equal  proportion of easy, medium, hard questions.  For model development and hyperparameter tuning, our development set consists of 282 examples, structured similarly  to the training set, with and equal proportion of easy, medium and hard questions.

Our primary evaluation data consists of 578 examples (the \concept{original} set) and 699 counterfactual examples. Additionally, to evaluate robustness to question complexity, we use a separate set of 732 easy, 507 medium and 719 hard questions that are distributionally similar to the original set.

\subsection{Models \& Baselines}
\label{sec:models-baselines}

We evaluate our approach of RL-based tuning with two open-weight models: SQLCoder 7B \cite{srivastava2023sqlcoder} and CodeGemma 7B~\cite{codegemmateam2024codegemmaopencodemodels}. Both models are specifically tuned for generating code, and RL-training them allows us to compare to specialized code-generation models.

We consider four families of baselines: 
\begin{inparaenum}[(a)]
\item The SQLCoder family, with 7B and 70B models, allows us to compare the RL-tuned SQLCoder against the original model, and also a much larger model that is trained in a similar fashion.
\item The \texttt{CodeGemma-7B} serves as an analogous baseline to the RL-tuned CodeGemma model.
\item \texttt{DeepSeek-R1-32B}~\cite{deepseekai2025deepseekr1incentivizingreasoningcapability}  derived from Qwen~\cite{hui2024qwen25codertechnicalreport} helps us compare to a larger and much more proficient RL-trained model.
\item \texttt{GPT-4o}~\cite{openai2024gpt4technicalreport} serves as an industrial strength closed weight point of comparison.
\end{inparaenum}

\subsection{Evaluation Metrics}
\label{sec:metrics}

Our primary evaluation of  the models from~\cref{sec:models-baselines} is the Exact Match Score (EMS), which measures the fraction of questions that were perfectly answered. Importantly, for questions that have multiple answers, EMS demands that they are all present in the system answer. The Relaxed Exact Match Score (REMS) provides partial credit and is defined as the normalized overlap between the predicted and ground truth answers. Finally, we also report the fraction of inputs for which the model generates syntactically invalid SQL queries. We measure this using the database engine as the fraction of instances where the engine reports an exception while executing the query.

\subsection{Training and Reward Details}
\label{sec:training-details}

We use the HuggingFace library~\cite{wolf-etal-2020-transformers} for our experiments. We use quantized models, and fine-tune with LORA adapters~\cite{dettmers2023qloraefficientfinetuningquantized, hu2021loralowrankadaptationlarge} with rank 16, scaling factor 32 and dropout rate 0.1 (all untuned). 
We trained for one epoch over the training data. We used the performance on the development set to pick the best model.
\Cref{tab:hyperparameters} and \cref{tab:quantization} in the appendix provide additional training details.

For the GRPO training loop, we generate $k=2$ candidate completions per input question using beam search, with a beam size 4. We set the value of $\beta$ in the loss \eqref{eq:loss} to $0.04$. Each completion is executed on a MariaDB database engine. 

Recall that our reward  consists of three aspects: syntactic correctness, partial correctness and exact match correctness. The syntactic correctness denoted as \( r_{\text{syn}} \), is assigned a value of 1 if the generated query executes without error. If the query fails to execute the model receives a fixed penalty of $r_{\text{err}}=-100$.
If the query executes successfully, the model receives a a partial correctness reward, \( r_{\text{partial}} \), based on the Relaxed Exact Match Score (REMS), scaled between 0 to 100. In addition, if the model's result exactly matches the gold answer, an exact match bonus, \( r_{\text{full}}=1000 \) is awarded.

\section{Experiments and Results}
\label{sec:results}

\begin{figure*}
    \centering
    \includegraphics[width=1.0\textwidth]{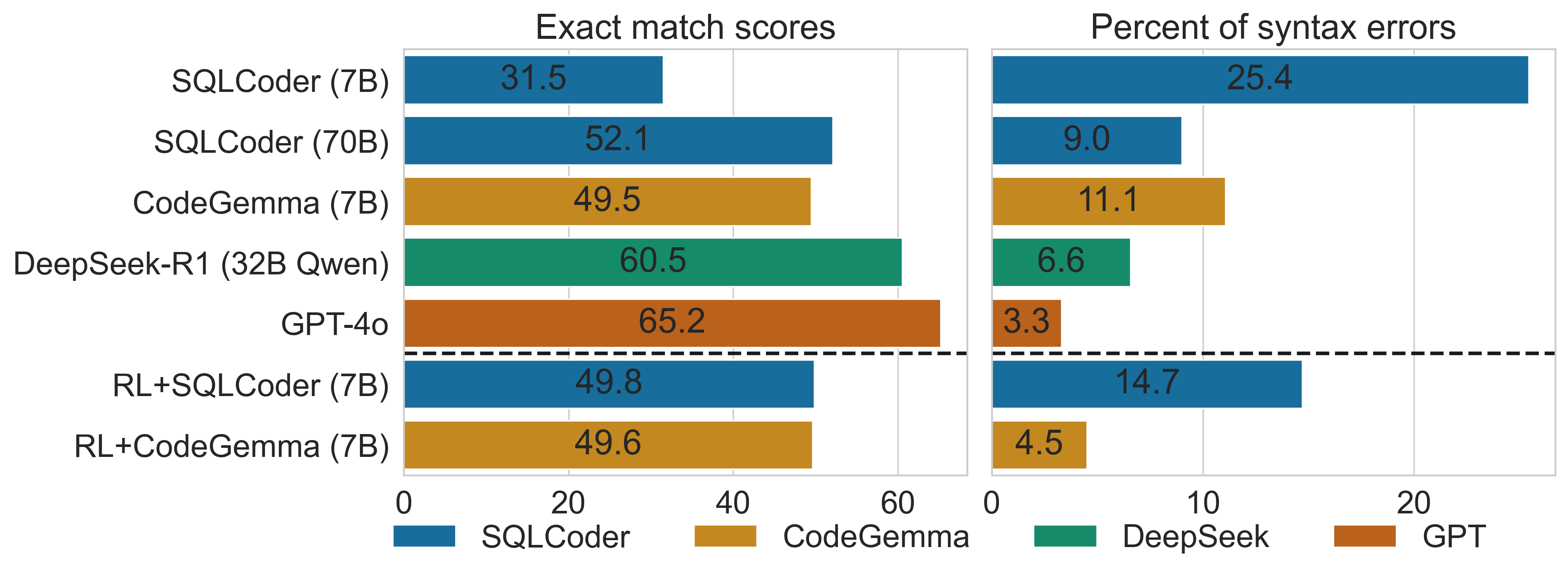}
    \caption{Exact Match Score and Error Percentage across multiple models on testing data. The colors in the figure represent model family. The two bars below the dashed line represent the two RL-tuned models.} 
    \label{fig:test_results}
\end{figure*}

In this section, we address the three research questions listed in~\cref{sec:intro}. Due to space constraints, we report the exact match scores and percentage of syntax errors here. The appendix presents the full set of results including the REMS.

\paragraph{Q1: Can RL-trained models outperform specialized SQL generation models in terms of answer quality and syntactic validity?}

\Cref{fig:test_results} compares the effectiveness of reinforcement learning (RL)-trained models against their untrained counterparts on the \concept{original} fraction of the test data. Specifically, the SQLCoder-7B (RL Tuned) model substantially outperforms its base variant in terms of the Exact Match Scores (EMS). The EMS of SQLCoder-7B improves from 31.49\% to 49.83\% with RL tuning. Moreover, RL tuning also reduces syntactic errors, lowering the error rate from 25.43\% to 14.71\% on the original set. This improved performance places the RL-tuned SQLCoder-7B model close to the larger SQL Coder-70B model (52.08\% EMS). 

We observe that while the more recent CodeGemma does not improve in terms of exact match (brown bars), the percent of syntax errors drops with RL tuning. Finally, our RL-tuned models underperform compared to DeepSeek-R1 and GPT-4o. This suggests that substantially more training (RL or otherwise) may lead to further gains.

\paragraph{Q2: How does RL training affect performance across queries of varying complexity?}

RL training enhances the model's performance across queries of varying complexity levels. Due to space constraints, the full results are in~\cref{tab:test_results} in the appendix; we present the highlights here. The SQLCoder-7B (RL Tuned) model demonstrates improvements over the baseline SQLCoder-7B model across Easy, Medium, and Hard query categories. On Easy queries, EMS improves from 60.79 to 72.27. Medium query performance increases from 33.73 to 51.08. Additionally, on Hard queries, the EMS doubles, rising from 13.27 to 27.53. This trend indicates that reinforcement learning particularly strengthens the model's capabilities in handling more challenging queries, reducing errors and improving accuracy systematically as query complexity increases.

\begin{table}

\centering
\begin{tabular}{l rr}
\toprule
Model & \shortstack{EMS} & \shortstack{Error Rate \% } \\
\midrule
SQLCoder-7B            & 29.18 & 28.04\\
SQL Coder 70B   & 47.93 & 10.73\\\midrule 
RL-SQLCoder-7B  & 47.07 & 16.74\\
\bottomrule
\end{tabular}

\caption{Counterfactual split performance for SQL-Coder models.}
\label{tab:q3_results}
\end{table}

\paragraph{Q3: Can RL training significantly enhance robustness on counterfactual temporal data?}

Robustness on counterfactual scenarios is enhanced by RL training. Comparing the performance of the base SQLCoder-7B and its RL-trained counterpart on counterfactual data reveals significant gains (See table~\ref{tab:q3_results}). The EMS score increases from 29.18 to 47.07, and the error rate decreases from 28.04\% to 16.74\%. This indicates a improvement in model robustness and adaptability to counterfactual perturbations. RL training therefore enhances the ability of the SQLCoder-7B model to reason accurately in counterfactual temporal contexts, aligning its performance closely with larger, more sophisticated models.

\section{Qualitative Analysis}
\label{sec:discussion}

In this section, we present a sample of responses that illustrate the impact of RL tuning on the outputs of SQLCoder. These examples are randomly chosen from the data.

\subsection{Inducing common sense}
\label{sec:common-sense}

We found several instances where the RL tuned models demonstrated better common sense understanding of the underlying domain than the original SQLCoder model. Consider the question below:

\begin{mdframed}
\footnotesize
\textbf{Question:}\\
\textit{How many years passed between Simon Fairweather's first and most recent International medal?}\\
\noindent
\textbf{Correct Answer:} 9

\end{mdframed}

\begin{figure}
\begin{mdframed}
\footnotesize
\textbf{SQLCoder-7B (Base) Response:}

\begin{lstlisting}[language=SQL]
SELECT MAX(m.year) - MIN(m.year) AS years_passed
FROM medal m
JOIN format f ON m.format_id = f.format_id
JOIN tournament t ON f.tournament_id = t.tournament_id
JOIN athlete a ON t.athlete_id = a.athlete_id
WHERE a.name = 'Simon Fairweather'
  AND t.name = 'International';
\end{lstlisting}
\small
\textbf{Output:} \texttt{NULL}
\end{mdframed}

\begin{mdframed}
\footnotesize

\noindent
\textbf{SQLCoder-7B (RL Tuned) Response:}

\begin{lstlisting}[language=SQL]
SELECT MAX(m.year) - MIN(m.year) AS years_passed
FROM medal m
JOIN format f ON m.format_id = f.format_id
JOIN tournament t ON f.tournament_id = t.tournament_id
JOIN athlete a ON t.athlete_id = a.athlete_id
WHERE a.name = 'Simon Fairweather';
\end{lstlisting}
\small
\textbf{Output:} 9
\end{mdframed}
\caption{SQL responses for example in \cref{sec:common-sense}.}
\label{fig:sql-common-sense}
\end{figure}

The outputs for this query are shown in \cref{fig:sql-common-sense}. The word `International' in the question refers to the level of the competition, such as the Olympic Games or World Championships, and  is not the name of any specific tournament. The RL-trained model learned not to filter for a tournament called "International," because no such tournament exists in the schema. Instead, it correctly constructed a query that considers all relevant tournaments where Simon Fairweather could have won an international medal. This demonstrates a practical improvement in the model's understanding of how international competition is represented in the data.

\subsection{Schema Adherence}
\label{sec:schema-adherence}
With RL tuning, the models learned to better adhere to the database schema. To illustrate this, consider the following question:
\begin{mdframed}
\footnotesize
\textbf{Question:} \\
\textit{What is the average number of medals won by Takaharu Furukawa in one year?}
\end{mdframed}

\begin{figure}

\begin{mdframed}
\footnotesize
\textbf{SQLCoder-7B (Base) Response:}

\begin{lstlisting}[language=SQL]
SELECT AVG(m.medal_count) AS average_medals
FROM (
  SELECT a.athlete_id, COUNT(m.medal_id) AS medal_count
  FROM medal m
  JOIN format f ON m.format_id = f.format_id
  JOIN tournament t ON f.tournament_id = t.tournament_id
  JOIN athlete a ON t.athlete_id = a.athlete_id
  WHERE a.name = 'Takaharu Furukawa'
  GROUP BY a.athlete_id
) AS subquery;
\end{lstlisting}
\small
\textbf{Output:} \\
\texttt{Error: Execution Failed: 1054 (42S22): Unknown column 'm.medal\_count' in 'SELECT'}
\end{mdframed}

\begin{mdframed}
\footnotesize
\textbf{SQLCoder-7B (RL Tuned) Response:}
\begin{lstlisting}[language=SQL]
SELECT AVG(COUNT) AS average_medals 
FROM (SELECT m.year, COUNT(*) AS COUNT FROM medal m 
JOIN format f ON m.format_id = f.format_id 
JOIN tournament t ON f.tournament_id = t.tournament_id 
JOIN athlete a ON t.athlete_id = a.athlete_id 
WHERE a.name = 'Takaharu Furukawa' GROUP BY m.year) AS subquery;
\end{lstlisting}
\small
\textbf{Output:} \\
1.33 (Correct Answer)
\end{mdframed}
\caption{SQL responses for example in \cref{sec:schema-adherence}.}
\label{fig:sql-schema-adherence}
\end{figure}

\Cref{fig:sql-schema-adherence} shows the SQL outputs by the untuned and tuned models. The schema does not contain any column called \texttt{medal\_count} in the \texttt{medal} table or any other table. Instead, to calculate how many medals an athlete has won, the query must use the SQL aggregation function \texttt{COUNT} on the \texttt{medal\_id} column. The base model’s error came from inventing a column name that does not exist, rather than using the correct SQL approach for counting rows.

Through reinforcement learning, the model reinforced its knowledge to use the \texttt{COUNT} function directly in the query to aggregate medal records, rather than referencing a non-existent column such as \texttt{medal\_count}. This adjustment made the model’s queries consistent with the schema and prevented errors related to unknown columns. As a result, the model avoided introducing invalid references and adhered to correct SQL aggregation practices and variable scope constraints. Such learning led to a reduction in schema-related errors and enabled the trained model to construct accurate queries for counting and aggregation tasks. It also tuned the model to consistently adhere to the schema as demonstrated in the improved result for the Charlotte Dujardin example.

\subsection{Failed Temporal Reasoning}
\label{sec:failed-temporal-reasoning}

Now, let us look at an example where, despite tuning with the SQL engine in the loop, the trained model does not produce the right query.

Consider the following question:

\begin{mdframed}
\footnotesize
\textbf{Question:} \\
\textit{How many medals did Valéria Kumizaki win in her twenties?}\\
\textbf{Correct Answer:} 5
\end{mdframed}

\begin{figure}
\noindent
\begin{mdframed}
\footnotesize
\textbf{SQLCoder-7B (RL Tuned) Response:}

\begin{lstlisting}[language=SQL]
SELECT COUNT(*) AS total_medals
FROM medal m
JOIN format f ON m.format_id = f.format_id
JOIN tournament t ON f.tournament_id = t.tournament_id
JOIN athlete a ON t.athlete_id = a.athlete_id
WHERE a.name = 'Valéria Kumizaki'
  AND m.year >= 2010
  AND m.year <= 2020;
\end{lstlisting}
\textbf{Output:} 9 (Incorrect)
\end{mdframed}

\begin{mdframed}
\footnotesize
\textbf{Deepseek-R1 (32B Qwen) Response:}

\begin{lstlisting}[language=SQL]
SELECT COUNT(m.medal_id) AS total_medals
FROM medal m
JOIN format f ON m.format_id = f.format_id
JOIN tournament t ON f.tournament_id = t.tournament_id
JOIN athlete a ON t.athlete_id = a.athlete_id
JOIN personalinformation p ON a.athlete_id = p.athlete_id
WHERE a.name = 'Valéria Kumizaki'
  AND m.year >= (p.birth_year + 20)
  AND m.year <= (p.birth_year + 29);
\end{lstlisting}
\textbf{Output:} 5 (Correct)
\end{mdframed}
\caption{SQL responses for example in~\cref{sec:failed-temporal-reasoning}.}
\label{fig:failed-temporal-reasoning}
\end{figure}

The outputs produced by the RL-tuned SQLCoder and Deepseek-R1 models are shown in~\cref{fig:failed-temporal-reasoning}.
The correct way to answer the question is to count medals won during the years in which the  age of the sportsperson was between 20 and 29, inclusive. The Deepseek-32B model response achieves this by joining the \texttt{personalinformation} table and calculating the relevant years using her \texttt{birth\_year} (\texttt{m.year >= (p.birth\_year + 20)} and \texttt{m.year <= (p.birth\_year + 29)}).

In contrast, the RL-tuned SQLCoder-7B model hard-coded the years 2010 to 2020 in the WHERE clause, rather than computing the correct age-based range using the schema. This approach is not schema-driven and does not adjust for the actual birth year of the athlete. As a result, it counted medals from years outside of Valéria Kumizaki’s twenties, returning 9 instead of the correct value 5.

This example demonstrates that while reinforcement learning reduced certain types of errors, such as those caused by referencing non-existent columns, it did not impart the necessary reasoning to compute age ranges based on birth year data from the schema. The Deepseek-32B model, in contrast, used a more better approach that combined correct schema usage with temporal reasoning. Such second-order reasoning, where the model connects date fields to age computations, remains a challenge for future work.

\subsection{Reward Hacking}
\label{sec:reward-hacking}

We observed a pattern of reward hacking behavior in CodeGemma during the GRPO iterations, which was not manifested with SQLCoder. The behavior emerged as a direct consequence of the Relaxed Exact Match Score ($r_{\mathrm{partial}}$) component of the reward function, which grants partial rewards for generating SQL queries that return partially correct answers. We found that CodeGemma produces SQL queries similar to the one in~\cref{fig:sql-reward-hacking}.

\begin{figure}
\begin{mdframed}
\footnotesize
\textbf{CodeGemma-7B Response:}
\begin{lstlisting}[language=SQL]
SELECT * FROM medal m 
JOIN format f ON m.format_id = f.format_id 
JOIN tournament t ON f.tournament_id = t.tournament_id 
JOIN athlete a ON t.athlete_id = a.athlete_id 
    WHERE a.name = 'Shevon Jemie Lai';
\end{lstlisting}
\textbf{Output:} All the medals won by the player along with the cities she participated in and the names of the tournament and their formats.
\end{mdframed}
\caption{Example of SQL used for reward hacking. See~\cref{sec:reward-hacking}}
\label{fig:sql-reward-hacking}
\end{figure}

While answering the above question instead of generating a query that identifies the city with the highest medal count (Kuala Lumpur) the model CodeGemma while RL training tried to maximize the partial reward and returned all the cities where Shevon Jemie Lai won medals. This behavior kept getting reinforced as outputing the same SQL above and just changing the name of the athlete allowed the model to gain a reward of 101 points. This example illustrates a risk of purely RL-driven efforts to train models to code.

\section{Discussion and Conclusion}
\label{sec:conclusion}

In this paper, we studied the question of whether we can improve an LLM that is already adept at producing SQL by having it interact with a database engine. We show that, using GRPO and simple deterministic reward functions, LLMs can generate accurate SQL queries, achieving improvement in both syntactic validity and logical correctness. Moreover, the improvements are especially marked on more difficult inputs.

Our work can be seen from the perspective of tool use, where LLMs can engage in structured reasoning tasks by using external tools. The examples and results presented in this paper illustrate how effectively LLMs can integrate symbolic reasoning and structured data querying when equipped with the right tools. Our experiments highlight the potential of LLMs to use low-level APIs, databases, and external systems like websearch as structured external environments. By employing reinforcement learning (RL),  models can learn to interact with these environments through structured language outputs (e.g., SQL queries), rather than through explicit supervised training on labeled examples.

\section*{Acknowledgments}
We thank the members of UtahNLP group for their valuable feedback. A.K. was supported by a University of Utah UROP fellowship over the course of this work. V.S. was supported in part by NSF awards \#2217154 and \#2411319.
\section*{Limitations}
\label{sec:limitations}

Our Reinforcement Learning approach is trained on academic-scale models. Larger models may behave differently under similar training protocols. This may restrict the generalization of our findings to significantly larger models that have much larger capacity.

Training with an SQL engine in the loop introduces latency. Unlike supervised fine-tuning, where gradients are computed directly from labeled examples, reinforcement learning with execution-based feedback involves repeated query generation and evaluation. This leads to slower training cycles.

The reward signal in our setup is solely based on the final answer. There is no direct supervision over the structure of the generated SQL. This can cause the model to generate SQL queries that are syntactically valid and answer-correct but logically flawed.

Weak supervision in the form of answer-only rewards may limit the models ability to learn intricate and nuanced concepts. When supervision is restricted to final execution outcomes, the model receives no guidance on intermediate steps or the query structure. As a result, it may struggle to acquire concepts that require multi-step logical inference or may require a large amount of time to improve its reasoning skills.

Reward hacking is also a concern. The model may learn to exploit artifacts in the reward computation to maximize return without generating accurate programs. This undermines the reliability and generalization of the model in unseen scenarios.

These limitations point to important directions for future work such as incorporating intermediate supervision, improving reward functions to mitigate reward hacking and scaling the method to larger models.

\bibliography{custom,anthology}

\begin{thebibliography}{21}
\providecommand{\natexlab}[1]{#1}

\bibitem[{Austin et~al.(2021)Austin, Odena, Nye, Bosma, Michalewski, Dohan, Jiang, Cai, Terry, Le, and Sutton}]{austin2021programsynthesislargelanguage}
Jacob Austin, Augustus Odena, Maxwell Nye, Maarten Bosma, Henryk Michalewski, David Dohan, Ellen Jiang, Carrie Cai, Michael Terry, Quoc Le, and Charles Sutton. 2021.
\newblock \href {https://arxiv.org/abs/2108.07732} {Program synthesis with large language models}.
\newblock \emph{Preprint}, arXiv:2108.07732.

\bibitem[{Berant et~al.(2013)Berant, Chou, Frostig, and Liang}]{berant-etal-2013-semantic}
Jonathan Berant, Andrew Chou, Roy Frostig, and Percy Liang. 2013.
\newblock \href {https://aclanthology.org/D13-1160/} {Semantic parsing on {F}reebase from question-answer pairs}.
\newblock In \emph{Proceedings of the 2013 Conference on Empirical Methods in Natural Language Processing}, pages 1533--1544, Seattle, Washington, USA. Association for Computational Linguistics.

\bibitem[{Brown et~al.(2020)Brown, Mann, Ryder, Subbiah, Kaplan, Dhariwal, Neelakantan, Shyam, Sastry, Askell, Agarwal, Herbert{-}Voss, Krueger, Henighan, Child, Ramesh, Ziegler, Wu, Winter, Hesse, Chen, Sigler, Litwin, Gray, Chess, Clark, Berner, McCandlish, Radford, Sutskever, and Amodei}]{DBLP:journals/corr/abs-2005-14165}
Tom~B. Brown, Benjamin Mann, Nick Ryder, Melanie Subbiah, Jared Kaplan, Prafulla Dhariwal, Arvind Neelakantan, Pranav Shyam, Girish Sastry, Amanda Askell, Sandhini Agarwal, Ariel Herbert{-}Voss, Gretchen Krueger, Tom Henighan, Rewon Child, Aditya Ramesh, Daniel~M. Ziegler, Jeffrey Wu, Clemens Winter, and 12 others. 2020.
\newblock \href {https://arxiv.org/abs/2005.14165} {Language models are few-shot learners}.
\newblock \emph{CoRR}, abs/2005.14165.

\bibitem[{Chen et~al.(2021{\natexlab{a}})Chen, Tworek, Jun, Yuan, de~Oliveira~Pinto, Kaplan, Edwards, Burda, Joseph, Brockman, Ray, Puri, Krueger, Petrov, Khlaaf, Sastry, Mishkin, Chan, Gray, Ryder, Pavlov, Power, Kaiser, Bavarian, Winter, Tillet, Such, Cummings, Plappert, Chantzis, Barnes, Herbert-Voss, Guss, Nichol, Paino, Tezak, Tang, Babuschkin, Balaji, Jain, Saunders, Hesse, Carr, Leike, Achiam, Misra, Morikawa, Radford, Knight, Brundage, Murati, Mayer, Welinder, McGrew, Amodei, McCandlish, Sutskever, and Zaremba}]{chen2021evaluatinglargelanguagemodels}
Mark Chen, Jerry Tworek, Heewoo Jun, Qiming Yuan, Henrique~Ponde de~Oliveira~Pinto, Jared Kaplan, Harri Edwards, Yuri Burda, Nicholas Joseph, Greg Brockman, Alex Ray, Raul Puri, Gretchen Krueger, Michael Petrov, Heidy Khlaaf, Girish Sastry, Pamela Mishkin, Brooke Chan, Scott Gray, and 39 others. 2021{\natexlab{a}}.
\newblock \href {https://arxiv.org/abs/2107.03374} {Evaluating large language models trained on code}.
\newblock \emph{Preprint}, arXiv:2107.03374.

\bibitem[{Chen et~al.(2021{\natexlab{b}})Chen, Tworek, Jun, Yuan, de~Oliveira~Pinto, Kaplan, Edwards, Burda, Joseph, Brockman, Ray, Puri, Krueger, Petrov, Khlaaf, Sastry, Mishkin, Chan, Gray, Ryder, Pavlov, Power, Kaiser, Bavarian, Winter, Tillet, Such, Cummings, Plappert, Chantzis, Barnes, Herbert{-}Voss, Guss, Nichol, Paino, Tezak, Tang, Babuschkin, Balaji, Jain, Saunders, Hesse, Carr, Leike, Achiam, Misra, Morikawa, Radford, Knight, Brundage, Murati, Mayer, Welinder, McGrew, Amodei, McCandlish, Sutskever, and Zaremba}]{DBLP:journals/corr/abs-2107-03374}
Mark Chen, Jerry Tworek, Heewoo Jun, Qiming Yuan, Henrique~Pond{\'{e}} de~Oliveira~Pinto, Jared Kaplan, Harri Edwards, Yuri Burda, Nicholas Joseph, Greg Brockman, Alex Ray, Raul Puri, Gretchen Krueger, Michael Petrov, Heidy Khlaaf, Girish Sastry, Pamela Mishkin, Brooke Chan, Scott Gray, and 39 others. 2021{\natexlab{b}}.
\newblock \href {https://arxiv.org/abs/2107.03374} {Evaluating large language models trained on code}.
\newblock \emph{CoRR}, abs/2107.03374.

\bibitem[{Clarke et~al.(2010)Clarke, Goldwasser, Chang, and Roth}]{clarke-etal-2010-driving}
James Clarke, Dan Goldwasser, Ming-Wei Chang, and Dan Roth. 2010.
\newblock \href {https://aclanthology.org/W10-2903/} {Driving semantic parsing from the world`s response}.
\newblock In \emph{Proceedings of the Fourteenth Conference on Computational Natural Language Learning}, pages 18--27, Uppsala, Sweden. Association for Computational Linguistics.

\bibitem[{DeepSeek-AI et~al.(2025)DeepSeek-AI, Guo, Yang, Zhang, Song, Zhang, Xu, Zhu, Ma, Wang, Bi, Zhang, Yu, Wu, Wu, Gou, Shao, Li, Gao, Liu, Xue, Wang, Wu, Feng, Lu, Zhao, Deng, Zhang, Ruan, Dai, Chen, Ji, Li, Lin, Dai, Luo, Hao, Chen, Li, Zhang, Bao, Xu, Wang, Ding, Xin, Gao, Qu, Li, Guo, Li, Wang, Chen, Yuan, Qiu, Li, Cai, Ni, Liang, Chen, Dong, Hu, Gao, Guan, Huang, Yu, Wang, Zhang, Zhao, Wang, Zhang, Xu, Xia, Zhang, Zhang, Tang, Li, Wang, Li, Tian, Huang, Zhang, Wang, Chen, Du, Ge, Zhang, Pan, Wang, Chen, Jin, Chen, Lu, Zhou, Chen, Ye, Wang, Yu, Zhou, Pan, Li, Zhou, Wu, Ye, Yun, Pei, Sun, Wang, Zeng, Zhao, Liu, Liang, Gao, Yu, Zhang, Xiao, An, Liu, Wang, Chen, Nie, Cheng, Liu, Xie, Liu, Yang, Li, Su, Lin, Li, Jin, Shen, Chen, Sun, Wang, Song, Zhou, Wang, Shan, Li, Wang, Wei, Zhang, Xu, Li, Zhao, Sun, Wang, Yu, Zhang, Shi, Xiong, He, Piao, Wang, Tan, Ma, Liu, Guo, Ou, Wang, Gong, Zou, He, Xiong, Luo, You, Liu, Zhou, Zhu, Xu, Huang, Li, Zheng, Zhu, Ma, Tang, Zha, Yan, Ren, Ren, Sha, Fu, Xu, Xie, Zhang,
  Hao, Ma, Yan, Wu, Gu, Zhu, Liu, Li, Xie, Song, Pan, Huang, Xu, Zhang, and Zhang}]{deepseekai2025deepseekr1incentivizingreasoningcapability}
DeepSeek-AI, Daya Guo, Dejian Yang, Haowei Zhang, Junxiao Song, Ruoyu Zhang, Runxin Xu, Qihao Zhu, Shirong Ma, Peiyi Wang, Xiao Bi, Xiaokang Zhang, Xingkai Yu, Yu~Wu, Z.~F. Wu, Zhibin Gou, Zhihong Shao, Zhuoshu Li, Ziyi Gao, and 181 others. 2025.
\newblock \href {https://arxiv.org/abs/2501.12948} {Deepseek-r1: Incentivizing reasoning capability in llms via reinforcement learning}.
\newblock \emph{Preprint}, arXiv:2501.12948.

\bibitem[{Dettmers et~al.(2023)Dettmers, Pagnoni, Holtzman, and Zettlemoyer}]{dettmers2023qloraefficientfinetuningquantized}
Tim Dettmers, Artidoro Pagnoni, Ari Holtzman, and Luke Zettlemoyer. 2023.
\newblock \href {https://arxiv.org/abs/2305.14314} {{PQLoRA: Efficient Finetuning of Quantized LLMs}}.
\newblock \emph{Preprint}, arXiv:2305.14314.

\bibitem[{Grattafiori et~al.(2024)Grattafiori, Dubey, Jauhri, Pandey, Kadian, Al-Dahle, Letman, Mathur, Schelten, Vaughan, Yang, Fan, Goyal, Hartshorn, Yang, Mitra, Sravankumar, Korenev, Hinsvark, Rao, Zhang, Rodriguez, Gregerson, Spataru, Roziere, Biron, Tang, Chern, Caucheteux, Nayak, Bi, Marra, McConnell, Keller, Touret, Wu, Wong, Ferrer, Nikolaidis, Allonsius, Song, Pintz, Livshits, Wyatt, Esiobu, Choudhary, Mahajan, Garcia-Olano, Perino, Hupkes, Lakomkin, AlBadawy, Lobanova, Dinan, Smith, Radenovic, Guzmán, Zhang, Synnaeve, Lee, Anderson, Thattai, Nail, Mialon, Pang, Cucurell, Nguyen, Korevaar, Xu, Touvron, Zarov, Ibarra, Kloumann, Misra, Evtimov, Zhang, Copet, Lee, Geffert, Vranes, Park, Mahadeokar, Shah, van~der Linde, Billock, Hong, Lee, Fu, Chi, Huang, Liu, Wang, Yu, Bitton, Spisak, Park, Rocca, Johnstun, Saxe, Jia, Alwala, Prasad, Upasani, Plawiak, Li, Heafield, Stone, El-Arini, Iyer, Malik, Chiu, Bhalla, Lakhotia, Rantala-Yeary, van~der Maaten, Chen, Tan, Jenkins, Martin, Madaan, Malo, Blecher,
  Landzaat, de~Oliveira, Muzzi, Pasupuleti, Singh, Paluri, Kardas, Tsimpoukelli, Oldham, Rita, Pavlova, Kambadur, Lewis, Si, Singh, Hassan, Goyal, Torabi, Bashlykov, Bogoychev, Chatterji, Zhang, Duchenne, Çelebi, Alrassy, Zhang, Li, Vasic, Weng, Bhargava, Dubal, Krishnan, Koura, Xu, He, Dong, Srinivasan, Ganapathy, Calderer, Cabral, Stojnic, Raileanu, Maheswari, Girdhar, Patel, Sauvestre, Polidoro, Sumbaly, Taylor, Silva, Hou, Wang, Hosseini, Chennabasappa, Singh, Bell, Kim, Edunov, Nie, Narang, Raparthy, Shen, Wan, Bhosale, Zhang, Vandenhende, Batra, Whitman, Sootla, Collot, Gururangan, Borodinsky, Herman, Fowler, Sheasha, Georgiou, Scialom, Speckbacher, Mihaylov, Xiao, Karn, Goswami, Gupta, Ramanathan, Kerkez, Gonguet, Do, Vogeti, Albiero, Petrovic, Chu, Xiong, Fu, Meers, Martinet, Wang, Wang, Tan, Xia, Xie, Jia, Wang, Goldschlag, Gaur, Babaei, Wen, Song, Zhang, Li, Mao, Coudert, Yan, Chen, Papakipos, Singh, Srivastava, Jain, Kelsey, Shajnfeld, Gangidi, Victoria, Goldstand, Menon, Sharma, Boesenberg,
  Baevski, Feinstein, Kallet, Sangani, Teo, Yunus, Lupu, Alvarado, Caples, Gu, Ho, Poulton, Ryan, Ramchandani, Dong, Franco, Goyal, Saraf, Chowdhury, Gabriel, Bharambe, Eisenman, Yazdan, James, Maurer, Leonhardi, Huang, Loyd, Paola, Paranjape, Liu, Wu, Ni, Hancock, Wasti, Spence, Stojkovic, Gamido, Montalvo, Parker, Burton, Mejia, Liu, Wang, Kim, Zhou, Hu, Chu, Cai, Tindal, Feichtenhofer, Gao, Civin, Beaty, Kreymer, Li, Adkins, Xu, Testuggine, David, Parikh, Liskovich, Foss, Wang, Le, Holland, Dowling, Jamil, Montgomery, Presani, Hahn, Wood, Le, Brinkman, Arcaute, Dunbar, Smothers, Sun, Kreuk, Tian, Kokkinos, Ozgenel, Caggioni, Kanayet, Seide, Florez, Schwarz, Badeer, Swee, Halpern, Herman, Sizov, Guangyi, Zhang, Lakshminarayanan, Inan, Shojanazeri, Zou, Wang, Zha, Habeeb, Rudolph, Suk, Aspegren, Goldman, Zhan, Damlaj, Molybog, Tufanov, Leontiadis, Veliche, Gat, Weissman, Geboski, Kohli, Lam, Asher, Gaya, Marcus, Tang, Chan, Zhen, Reizenstein, Teboul, Zhong, Jin, Yang, Cummings, Carvill, Shepard, McPhie,
  Torres, Ginsburg, Wang, Wu, U, Saxena, Khandelwal, Zand, Matosich, Veeraraghavan, Michelena, Li, Jagadeesh, Huang, Chawla, Huang, Chen, Garg, A, Silva, Bell, Zhang, Guo, Yu, Moshkovich, Wehrstedt, Khabsa, Avalani, Bhatt, Mankus, Hasson, Lennie, Reso, Groshev, Naumov, Lathi, Keneally, Liu, Seltzer, Valko, Restrepo, Patel, Vyatskov, Samvelyan, Clark, Macey, Wang, Hermoso, Metanat, Rastegari, Bansal, Santhanam, Parks, White, Bawa, Singhal, Egebo, Usunier, Mehta, Laptev, Dong, Cheng, Chernoguz, Hart, Salpekar, Kalinli, Kent, Parekh, Saab, Balaji, Rittner, Bontrager, Roux, Dollar, Zvyagina, Ratanchandani, Yuvraj, Liang, Alao, Rodriguez, Ayub, Murthy, Nayani, Mitra, Parthasarathy, Li, Hogan, Battey, Wang, Howes, Rinott, Mehta, Siby, Bondu, Datta, Chugh, Hunt, Dhillon, Sidorov, Pan, Mahajan, Verma, Yamamoto, Ramaswamy, Lindsay, Lindsay, Feng, Lin, Zha, Patil, Shankar, Zhang, Zhang, Wang, Agarwal, Sajuyigbe, Chintala, Max, Chen, Kehoe, Satterfield, Govindaprasad, Gupta, Deng, Cho, Virk, Subramanian, Choudhury,
  Goldman, Remez, Glaser, Best, Koehler, Robinson, Li, Zhang, Matthews, Chou, Shaked, Vontimitta, Ajayi, Montanez, Mohan, Kumar, Mangla, Ionescu, Poenaru, Mihailescu, Ivanov, Li, Wang, Jiang, Bouaziz, Constable, Tang, Wu, Wang, Wu, Gao, Kleinman, Chen, Hu, Jia, Qi, Li, Zhang, Zhang, Adi, Nam, Yu, Wang, Zhao, Hao, Qian, Li, He, Rait, DeVito, Rosnbrick, Wen, Yang, Zhao, and Ma}]{grattafiori2024llama3herdmodels}
Aaron Grattafiori, Abhimanyu Dubey, Abhinav Jauhri, Abhinav Pandey, Abhishek Kadian, Ahmad Al-Dahle, Aiesha Letman, Akhil Mathur, Alan Schelten, Alex Vaughan, Amy Yang, Angela Fan, Anirudh Goyal, Anthony Hartshorn, Aobo Yang, Archi Mitra, Archie Sravankumar, Artem Korenev, Arthur Hinsvark, and 542 others. 2024.
\newblock \href {https://arxiv.org/abs/2407.21783} {The llama 3 herd of models}.
\newblock \emph{Preprint}, arXiv:2407.21783.

\bibitem[{Hu et~al.(2021)Hu, Shen, Wallis, Allen-Zhu, Li, Wang, Wang, and Chen}]{hu2021loralowrankadaptationlarge}
Edward~J. Hu, Yelong Shen, Phillip Wallis, Zeyuan Allen-Zhu, Yuanzhi Li, Shean Wang, Lu~Wang, and Weizhu Chen. 2021.
\newblock \href {https://arxiv.org/abs/2106.09685} {{LoRA: Low-Rank Adaptation of Large Language Models}}.
\newblock \emph{Preprint}, arXiv:2106.09685.

\bibitem[{Hui et~al.(2024)Hui, Yang, Cui, Yang, Liu, Zhang, Liu, Zhang, Yu, Lu, Dang, Fan, Zhang, Yang, Men, Huang, Zheng, Miao, Quan, Feng, Ren, Ren, Zhou, and Lin}]{hui2024qwen25codertechnicalreport}
Binyuan Hui, Jian Yang, Zeyu Cui, Jiaxi Yang, Dayiheng Liu, Lei Zhang, Tianyu Liu, Jiajun Zhang, Bowen Yu, Keming Lu, Kai Dang, Yang Fan, Yichang Zhang, An~Yang, Rui Men, Fei Huang, Bo~Zheng, Yibo Miao, Shanghaoran Quan, and 5 others. 2024.
\newblock \href {https://arxiv.org/abs/2409.12186} {Qwen2.5-coder technical report}.
\newblock \emph{Preprint}, arXiv:2409.12186.

\bibitem[{Kulkarni et~al.(2025)Kulkarni, Dixit, Srikumar, Roth, and Gupta}]{kulkarni-etal-2025-llm}
Atharv Kulkarni, Kushagra Dixit, Vivek Srikumar, Dan Roth, and Vivek Gupta. 2025.
\newblock Llm-symbolic integration for robust temporal tabular reasoning.
\newblock In \emph{Findings of the Association for Computational Linguistics: ACL 2025}, Vienna, Austria. Association for Computational Linguistics.

\bibitem[{Liang et~al.(2011)Liang, Jordan, and Klein}]{liang-etal-2011-learning}
Percy Liang, Michael Jordan, and Dan Klein. 2011.
\newblock \href {https://aclanthology.org/P11-1060/} {Learning dependency-based compositional semantics}.
\newblock In \emph{Proceedings of the 49th Annual Meeting of the Association for Computational Linguistics: Human Language Technologies}, pages 590--599, Portland, Oregon, USA. Association for Computational Linguistics.

\bibitem[{OpenAI et~al.(2024)OpenAI, Achiam, Adler, Agarwal, Ahmad, Akkaya, Aleman, Almeida, Altenschmidt, Altman, Anadkat, Avila, Babuschkin, Balaji, Balcom, Baltescu, Bao, Bavarian, Belgum, Bello, Berdine, Bernadett-Shapiro, Berner, Bogdonoff, Boiko, Boyd, Brakman, Brockman, Brooks, Brundage, Button, Cai, Campbell, Cann, Carey, Carlson, Carmichael, Chan, Chang, Chantzis, Chen, Chen, Chen, Chen, Chen, Chess, Cho, Chu, Chung, Cummings, Currier, Dai, Decareaux, Degry, Deutsch, Deville, Dhar, Dohan, Dowling, Dunning, Ecoffet, Eleti, Eloundou, Farhi, Fedus, Felix, Fishman, Forte, Fulford, Gao, Georges, Gibson, Goel, Gogineni, Goh, Gontijo-Lopes, Gordon, Grafstein, Gray, Greene, Gross, Gu, Guo, Hallacy, Han, Harris, He, Heaton, Heidecke, Hesse, Hickey, Hickey, Hoeschele, Houghton, Hsu, Hu, Hu, Huizinga, Jain, Jain, Jang, Jiang, Jiang, Jin, Jin, Jomoto, Jonn, Jun, Kaftan, Łukasz Kaiser, Kamali, Kanitscheider, Keskar, Khan, Kilpatrick, Kim, Kim, Kim, Kirchner, Kiros, Knight, Kokotajlo, Łukasz Kondraciuk,
  Kondrich, Konstantinidis, Kosic, Krueger, Kuo, Lampe, Lan, Lee, Leike, Leung, Levy, Li, Lim, Lin, Lin, Litwin, Lopez, Lowe, Lue, Makanju, Malfacini, Manning, Markov, Markovski, Martin, Mayer, Mayne, McGrew, McKinney, McLeavey, McMillan, McNeil, Medina, Mehta, Menick, Metz, Mishchenko, Mishkin, Monaco, Morikawa, Mossing, Mu, Murati, Murk, Mély, Nair, Nakano, Nayak, Neelakantan, Ngo, Noh, Ouyang, O'Keefe, Pachocki, Paino, Palermo, Pantuliano, Parascandolo, Parish, Parparita, Passos, Pavlov, Peng, Perelman, de~Avila Belbute~Peres, Petrov, de~Oliveira~Pinto, Michael, Pokorny, Pokrass, Pong, Powell, Power, Power, Proehl, Puri, Radford, Rae, Ramesh, Raymond, Real, Rimbach, Ross, Rotsted, Roussez, Ryder, Saltarelli, Sanders, Santurkar, Sastry, Schmidt, Schnurr, Schulman, Selsam, Sheppard, Sherbakov, Shieh, Shoker, Shyam, Sidor, Sigler, Simens, Sitkin, Slama, Sohl, Sokolowsky, Song, Staudacher, Such, Summers, Sutskever, Tang, Tezak, Thompson, Tillet, Tootoonchian, Tseng, Tuggle, Turley, Tworek, Uribe, Vallone,
  Vijayvergiya, Voss, Wainwright, Wang, Wang, Wang, Ward, Wei, Weinmann, Welihinda, Welinder, Weng, Weng, Wiethoff, Willner, Winter, Wolrich, Wong, Workman, Wu, Wu, Wu, Xiao, Xu, Yoo, Yu, Yuan, Zaremba, Zellers, Zhang, Zhang, Zhao, Zheng, Zhuang, Zhuk, and Zoph}]{openai2024gpt4technicalreport}
OpenAI, Josh Achiam, Steven Adler, Sandhini Agarwal, Lama Ahmad, Ilge Akkaya, Florencia~Leoni Aleman, Diogo Almeida, Janko Altenschmidt, Sam Altman, Shyamal Anadkat, Red Avila, Igor Babuschkin, Suchir Balaji, Valerie Balcom, Paul Baltescu, Haiming Bao, Mohammad Bavarian, Jeff Belgum, and 262 others. 2024.
\newblock \href {https://arxiv.org/abs/2303.08774} {Gpt-4 technical report}.
\newblock \emph{Preprint}, arXiv:2303.08774.

\bibitem[{Schulman et~al.(2017)Schulman, Wolski, Dhariwal, Radford, and Klimov}]{schulman2017proximalpolicyoptimizationalgorithms}
John Schulman, Filip Wolski, Prafulla Dhariwal, Alec Radford, and Oleg Klimov. 2017.
\newblock \href {https://arxiv.org/abs/1707.06347} {Proximal policy optimization algorithms}.
\newblock \emph{Preprint}, arXiv:1707.06347.

\bibitem[{Shao et~al.(2024)Shao, Wang, Zhu, Xu, Song, Bi, Zhang, Zhang, Li, Wu, and Guo}]{shao2024deepseekmathpushinglimitsmathematical}
Zhihong Shao, Peiyi Wang, Qihao Zhu, Runxin Xu, Junxiao Song, Xiao Bi, Haowei Zhang, Mingchuan Zhang, Y.~K. Li, Y.~Wu, and Daya Guo. 2024.
\newblock \href {https://arxiv.org/abs/2402.03300} {Deepseekmath: Pushing the limits of mathematical reasoning in open language models}.
\newblock \emph{Preprint}, arXiv:2402.03300.

\bibitem[{Srivastava and Aw(2023)}]{srivastava2023sqlcoder}
Rishabh Srivastava and Wendy Aw. 2023.
\newblock Open-sourcing sqlcoder: A state-of-the-art llm for sql generation.
\newblock \url{https://defog.ai/blog/open-sourcing-sqlcoder}.
\newblock Accessed: 2025-05-19.

\bibitem[{Team et~al.(2024)Team, Zhao, Hui, Howland, Nguyen, Zuo, Hu, Choquette-Choo, Shen, Kelley, Bansal, Vilnis, Wirth, Michel, Choy, Joshi, Kumar, Hashmi, Agrawal, Gong, Fine, Warkentin, Hartman, Ni, Korevec, Schaefer, and Huffman}]{codegemmateam2024codegemmaopencodemodels}
CodeGemma Team, Heri Zhao, Jeffrey Hui, Joshua Howland, Nam Nguyen, Siqi Zuo, Andrea Hu, Christopher~A. Choquette-Choo, Jingyue Shen, Joe Kelley, Kshitij Bansal, Luke Vilnis, Mateo Wirth, Paul Michel, Peter Choy, Pratik Joshi, Ravin Kumar, Sarmad Hashmi, Shubham Agrawal, and 8 others. 2024.
\newblock \href {https://arxiv.org/abs/2406.11409} {Codegemma: Open code models based on gemma}.
\newblock \emph{Preprint}, arXiv:2406.11409.

\bibitem[{Wang et~al.(2024)Wang, Chen, Yuan, Zhang, Li, Peng, and Ji}]{wang2024executablecodeactionselicit}
Xingyao Wang, Yangyi Chen, Lifan Yuan, Yizhe Zhang, Yunzhu Li, Hao Peng, and Heng Ji. 2024.
\newblock \href {https://arxiv.org/abs/2402.01030} {Executable code actions elicit better llm agents}.
\newblock \emph{Preprint}, arXiv:2402.01030.

\bibitem[{Wolf et~al.(2020)Wolf, Debut, Sanh, Chaumond, Delangue, Moi, Cistac, Rault, Louf, Funtowicz, Davison, Shleifer, von Platen, Ma, Jernite, Plu, Xu, Le~Scao, Gugger, Drame, Lhoest, and Rush}]{wolf-etal-2020-transformers}
Thomas Wolf, Lysandre Debut, Victor Sanh, Julien Chaumond, Clement Delangue, Anthony Moi, Pierric Cistac, Tim Rault, Remi Louf, Morgan Funtowicz, Joe Davison, Sam Shleifer, Patrick von Platen, Clara Ma, Yacine Jernite, Julien Plu, Canwen Xu, Teven Le~Scao, Sylvain Gugger, and 3 others. 2020.
\newblock \href {https://doi.org/10.18653/v1/2020.emnlp-demos.6} {Transformers: State-of-the-art natural language processing}.
\newblock In \emph{Proceedings of the 2020 Conference on Empirical Methods in Natural Language Processing: System Demonstrations}, pages 38--45, Online. Association for Computational Linguistics.

\bibitem[{Zhou et~al.(2024)Zhou, Xu, Zhu, Zhou, Lo, Sridhar, Cheng, Ou, Bisk, Fried, Alon, and Neubig}]{zhou2024webarenarealisticwebenvironment}
Shuyan Zhou, Frank~F. Xu, Hao Zhu, Xuhui Zhou, Robert Lo, Abishek Sridhar, Xianyi Cheng, Tianyue Ou, Yonatan Bisk, Daniel Fried, Uri Alon, and Graham Neubig. 2024.
\newblock \href {https://arxiv.org/abs/2307.13854} {{WebArena: A Realistic Web Environment for Building Autonomous Agents}}.
\newblock \emph{Preprint}, arXiv:2307.13854.

\end{thebibliography}
\bibliographystyle{acl_natbib}

\appendix

\section{Appendix}
\label{sec:appendix}

\subsection{Compute Environment}
For training we used a compute node with eight Tesla V100-SXM2-16GB  running in  parallel with NVLink. For running the baselines across all models we used 1/2 A100 GPU with 6 cores and 128 GB of memory per job. 

\subsection{Hyperparameters}

\begin{table}[h]
\centering
\begin{tabular}{ll}
\toprule
\textbf{Hyperparameter}            & \textbf{Value}  \\
\midrule
Learning rate                      & \( 10^{-5}\)    \\
Batch size                         & 2 prompts       \\
Number of epochs                   & 1               \\
% Checkpoint interval              & 300 completions \\
Beam width                         & 4               \\
Max prompt length                  & 3000 tokens     \\
Max completion length              & 256 tokens      \\
KL penalty coefficient (\(\beta\)) & 0.04            \\
\bottomrule
\end{tabular}
\caption{Training Hyperparameters.}
\label{tab:hyperparameters}
\end{table}

\Cref{tab:hyperparameters} displays the hyperparameters used while training both the SQLCoder-7B and CodeGemma-7B models using Reinforcement Learning.

\subsection{Quantization}

\begin{table}[ht]
  \centering
  \begin{tabular}{ll}
    \toprule
    Model                               & Quantization \\
    \midrule
    SQL Coder 70 B                      & 4 bit        \\
    SQL Coder 7 B                       & 8 bit        \\
    SQL Coder 7 B (RL Tuned)            & 8 bit        \\
    Code Gemma 7 B                      & 16 bit       \\
    Code Gemma 7 B (RL Tuned)           & 16 bit       \\
    Deep Seek 32 B Qwen 2.5             & 16 bit       \\
    \bottomrule
  \end{tabular}
  \caption{Quantization of models}
  \label{tab:quantization}
\end{table}

\Cref{tab:quantization} shows the quantizations used for each of the models while evaluating the baselines and while training.

\subsection{Results in Tabular format}
\begin{table*}[ht]
\small
\centering
\begin{tabular}{l l c c c c c}
\toprule
\textbf{Model name} & \textbf{Metric} & \shortstack{\textbf{Original} \\ \textbf{(578)}} & \shortstack{\textbf{Counterfactual} \\ \textbf{(699)}} & \shortstack{\textbf{Easy} \\ \textbf{(732)}} & \shortstack{\textbf{Medium} \\ \textbf{(507)}} & \shortstack{\textbf{Hard} \\ \textbf{(719)}} \\
\midrule
\multirow{2}{*}{GPT-4o} & REMS & 66.76 & 62.40 & 81.27 & 75.88 & 64.38  \\
                        & EMS & 65.22 & 60.94 & 78.89 & 75.15 & 64.31 \\
                        & \#Error & 19 (3.29\%) & 18 (2.58\%) & 3 (0.41\%) & 10 (1.97\%) & 16 (2.23\%) \\
                        
\midrule
\multirow{2}{*}{\shortstack{DeepSeek-R1 \\ 32B-Qwen}} 
                & REMS & 61.02 & 64.79 & 79.7 & 62.08 & 55.11 \\
                & EMS & 60.55 & 63.73 & 79.51 & 61.74 & 53.85 \\
                & \#Error & 38 (6.57\%) & 27 (3.86\%) & 2 (0.27\%) & 39 (7.69\%) & 37 (5.15\%) \\

\midrule
\multirow{2}{*}{\shortstack{SQL Coder 70B\\ (4 bit)}} 
& REMS & 53.37 & 49.32 & 77.39 & 54.05 & 30.19 \\
& EMS & 52.08 & 47.93 & 77.39 & 51.19 & 28.91 \\
 & \#Error & 52 (9.00\%) & 75 (10.73\%) & 2 (0.27\%) & 0 (0.00\%) & 50 (6.95\%) \\

\midrule
\multirow{2}{*}{CodeGemma-7B} & REMS & 50.21 & 43.60 & 71.64 & 63.51 & 34.63 \\
                            & EMS & 49.48 & 42.63 & 71.45 & 62.92 & 32.34 \\
                             & \#Error & 64 (11.07\%) & 100 (14.31\%) & 90 (12.30\%) & 2 (0.39\%) & 37 (5.15\%) \\
                             
\midrule
\multirow{2}{*}{SQLCoder-7B}  & REMS & 32.32 & 30.00 & 60.99 & 36.21 & 13.27 \\
& EMS & 31.49 & 29.18 & 60.79 & 33.73 & 13.27 \\
 & \#Error & 147 (25.43\%) & 196 (28.04\%) & 49 (6.69\%) & 126 (24.85\%) & 142 (19.75\%) \\

\midrule
\midrule
\multirow{2}{*}{\shortstack{SQLCoder-7B \\ (RL Tuned)}} 
& REMS & 51.10 & 48.36 & 72.36 & 54.4 & 28.63 \\
& EMS & 49.83 & 47.07 & 72.27 & 51.08 & 27.53 \\
 & \#Error & 85 (14.71\%) & 117 (16.74\%) & 81 (11.07\%) & 41 (8.09\%) & 108 (15.02\%) \\
\midrule
\multirow{2}{*}{\shortstack{CodeGemma-7B \\ (RL Tuned)}}
& REMS & 50.38 & 43.31 & 72.6 & 61.54 & 35.87 \\
& EMS & 49.65 & 42.20 & 72.40 & 60.95 & 33.17 \\
& \#Error & 26 (4.50\%) & 52 (7.44\%) & 32 (4.37\%) & 2 (0.39\%) & 18 (2.50\%) \\
\bottomrule
\end{tabular}
\caption{Test Results }
\label{tab:test_results}
\end{table*}

\Cref{tab:test_results} describes the results of our baseline experiments and the performance of the trained models on the test set of TEMPTABQA-C.

\paragraph{GPT-4o}:
The model GPT-4o serves as a baseline with the highest performance across all the subsets. Its EMS scores are 65.22 on Original, 60.94 on CounterFactual, 78.89 on Easy, 75.15 on Medium, 64.31 on Hard. Error percentages are relatively low: 3.29\% on Original, 2.588\% on CounterFactual, 0.41\% on Easy, 1.97\% on Medium, 2.23\% on Hard. Compared to all models GPT-4o achieves the highest EMS and the lowest error percentages consistently across subsets.

\paragraph{DeepSeek-R1 (32B-Qwen)}:
DeepSeek achieves an EMS of 60.55 on Original, 63.73 on CounterFactual, 79.51 on Easy, 61.74 on Medium and 53.85 on Hard. Deep Seek-R1 performs second best on Original, Counterfactual, Easy and Hard subsets but falls slightly behind SQL-Coder 70B on Medium.

\paragraph{SQLCoder-70B}:
SQLCoder -70B with 4 bit quantization achieves 52.08 on Original, 47.93 on CounterFactual, 77.39 on Easy, 51.19 on Medium and 28.91 on Hard. SQL-Coder-70B trails GPT-4o and DeepSeek-R1 on most subsets however it achieves very less errors on Easy and Medium subsets.

\paragraph{CodeGemma 7B}: CodeGemma-7B achieves 49.48 on Original, 42.63 on CounterFactual, 71.45 on Easy, 62.92 on Medium, and 32.34 on Hard. CodeGemma-7B although being a 7B parameter model on the original dataset it performs very close to the much larger SQLCoder-70B model. It also performs better than the SQLCoder-7B model.

\paragraph{SQLCoder 7B}: SQLCoder-7B achieves 31.49 on Original, 29.18 on CounterFactual, 60.79 on Easy, 33.73 on Medium, and 13.27 on Hard. This model without reinforcement learning shows the largest error rates and falls behind all the models.

\paragraph{SQLCoder-7B RL Tuned}:  With Reinforcement Learning SQLCoder 7B shows significant improvement over its original version. SQLCOder-7B scores 49.83 on Original, 47.07 on CounterFactual, 72.27 on Easy, 51.08 on Medium and 27.53 on  Hard. While still much behind GPT-4o and DeepSeek-R1, the RL tuning places SQLCoder-7B very close to the much larger SQLCoder70B model.

\paragraph{CodeGemma-7B RL Tuned}: Reinforcement Learning presents almost no score improvements to CodeGemma-7B, however significant drop in error rates happen to CodeGemma-7B after RL tuning. Error percentages decrease to 4.5\% on Original, 7.44\% on CounterFactual, 4.37\% on Easy, and 0.39\% on Medium and 2.5\% on Hard after RL tuning.

\end{document}